\documentclass[10pt,twocolumn,letterpaper]{article}

\usepackage{iccv}
\usepackage{times}
\usepackage{epsfig}
\usepackage{graphicx}
\usepackage{amsmath}
\usepackage{amssymb}

\usepackage{times}
\usepackage{epsfig}
\usepackage{graphicx}
\usepackage{amsmath}
\usepackage{amssymb}
\usepackage{nccmath}

\usepackage{dsfont}
\usepackage{siunitx}
\usepackage{subfigure}
\usepackage{multirow}
\usepackage{caption}
\usepackage{booktabs}       %
\usepackage{standalone}
\usepackage{xspace}
\usepackage{comment}
\usepackage{multirow}
\usepackage{bbm}
\usepackage{xcolor}

\makeatletter
\DeclareRobustCommand\onedot{\futurelet\@let@token\@onedot}
\def\@onedot{\ifx\@let@token.\else.\null\fi\xspace}

\makeatother

\newcommand{\myparagraph}[1]{\vspace{5pt}\noindent{\bf #1}}

\newcommand{\TASKNAME}{Change Captioning\xspace}
\newcommand{\DATASETNAME}{CLEVR-Change Dataset\xspace}
\newcommand{\DATASETABBR}{CLEVR-Change\xspace}
\newcommand{\MODELNAME}{Dual Dynamic Attention Model\xspace}
\newcommand{\MODELABBR}{DUDA\xspace} %

\usepackage[pagebackref=true,breaklinks=true,letterpaper=true,colorlinks,bookmarks=false]{hyperref}

\iccvfinalcopy %

\def\iccvPaperID{5135} %

\begin{document}

\title{Robust Change Captioning}

\author{Dong Huk Park, Trevor Darrell, Anna Rohrbach \vspace{3mm} \\
University of California, Berkeley}

\maketitle
\begin{abstract}
Describing what has changed in a scene can be useful to a user, but only if generated text focuses on what is semantically relevant. It is thus important to distinguish distractors (e.g. a viewpoint change) from relevant changes (e.g. an object has moved). 
We present a novel \MODELNAME{} (\MODELABBR{}) to perform robust \TASKNAME{}. Our model learns to distinguish distractors from semantic changes, localize the changes via Dual Attention over ``before'' and ``after'' images, and accurately describe them in natural language via Dynamic Speaker, by adaptively focusing on the necessary visual inputs (e.g. ``before'' or ``after'' image). 
To study the problem in depth, we collect a \DATASETABBR{} dataset, built off the CLEVR engine, with 5 types of scene changes. We benchmark a number of baselines on our dataset, and systematically study different change types and robustness to distractors. We show the superiority of our \MODELABBR{} model in terms of both change captioning and localization. We also show that our approach is general, obtaining state-of-the-art results on the recent realistic Spot-the-Diff dataset which has no distractors.
\end{abstract}

\section{\label{sec:intro} Introduction}

\begin{figure}[t]
    \centering
    \includegraphics[width=\linewidth]{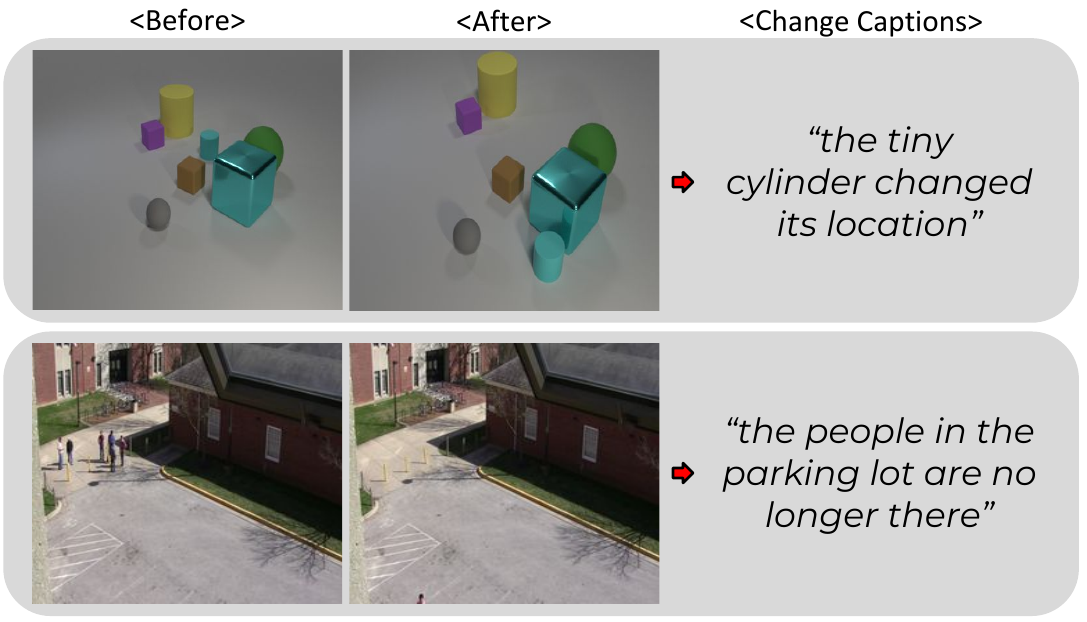}
    \caption{Robust \TASKNAME{} requires semantic visual understanding in which scene change must be distinguished from mere viewpoint shift (top row). Not only does it require accurate localization of a change, but it also requires communicating the change in natural language. Our \MODELNAME{} (\MODELABBR{}) demonstrates such capacity via a specialized attention mechanism.%
    }
    \label{fig:teaser}
\end{figure}

We live in a dynamic world where things change all the time. 
Change detection in images is a long-standing research problem, with applications in a variety of domains such as facility monitoring,  medical imaging, and aerial photography  \cite{gueguen2015large,patriarche2004review,sakurada2015change}. A key challenge in change detection is to distinguish the relevant changes from the irrelevant ones \cite{radke2005image} since the former are those that should likely trigger a notification. 
Existing systems aim to sense or localize a change, but typically do not convey  detailed semantic content. This is an important limitation for a realistic application, where analysts would benefit from such knowledge, helping them to better understand and judge the significance of the change. Alerting a user on every detected difference likely will lead to a frustrated operator; moreover, it is desirable to have a change detection system that does not output a binary indicator of change/no-change, but instead outputs a concise description of what has changed, and where.

Expressing image content in natural language is an active area of Artificial Intelligence research, with numerous approaches to image captioning having been recently proposed \cite{anderson2018bottom,donahue15cvpr,lu2018neural,xu2015show}. These methods have the benefit of conveying visual content to human users in a concise and natural way. They can be especially useful, when tailored to a specific task or objective, such as e.g. explaining the model's predictions \cite{hendricks2018grounding,park2018multimodal} or generating non-ambiguous referring expressions for specific image regions \cite{luo2017comprehension,yu2017joint}.

In this work we investigate robust \emph{\TASKNAME{}}, where an important scene change has to be identified and conveyed using natural language in the presence of distractors (where only an illumination or viewpoint change occurred). We aim to generate detailed and informative descriptions that  refer to the changed objects in complex scenes (see \autoref{fig:teaser}).

To distinguish an irrelevant distractor from an actual change (e.g. an object moved), one needs to ``compare'' the two images and find correspondences and disagreements. %
We propose a \emph{\MODELNAME{} (\MODELABBR{})} that learns to localize the changes via a specialized attention mechanism. It consists of two components: \emph{Dual Attention} that predicts a separate spatial attention for each image in the ``before''/``after'' pair, and a \emph{Dynamic Speaker} that generates a change description by semantically modulating focus among the visual features relayed from the Dual Attention. Both components are neural networks that are trained jointly with only caption-level supervision, i.e. no information about the change location is used during training.

In order to study \TASKNAME{} in the presence of distractors, we build a \emph{\DATASETNAME{}}. We rely on the image generation engine by \cite{johnson2017clevr}, which allows us to produce complex compositional scenes. We create pairs of ``before'' and ``after'' images with: (a) only illumination/viewpoint change (distractors), and (b) illumination/viewpoint change combined with a scene change. We consider 5 scene change types (color/material change, adding/dropping/moving an object), and collect almost 80K image pairs. We augment the image pairs with automatically generated change captions (see \autoref{fig:dataset_examples}). Note that in the recently proposed  Spot-the-Diff dataset~\cite{jhamtani2018learning}, the task also is to generate change captions for a pair of images. However, their problem statement is different from ours in that: 1) they assume a change in each image pair while our goal is to be robust to distractors, 2) the images are aligned (no viewpoint shift), 3) change localization can not be evaluated as ground-truth is not available in \cite{jhamtani2018learning}.

We first evaluate our novel \MODELABBR{} model on the \DATASETABBR{} dataset, and compare it to a number of baselines, including a naive pixel-difference captioning baseline. We show that our approach outperforms the baselines in terms of change caption correctness as well as change localization. The most challenging change types to describe are object movement and texture change, while movement is also the hardest to localize. 
We also show that our approach is general, applying it to the Spot-the-Diff dataset~\cite{jhamtani2018learning}. Given the same visual inputs as \cite{jhamtani2018learning}, our model matches or outperforms their approach. 

\section{\label{sec:related} Related Work}

Here we discuss prior work on change detection, task-specific image captioning, and attention mechanism.

\myparagraph{Change detection}
One popular domain for image-based change detection is aerial imagery \cite{liu2018change,tian2014building,zanetti2016generalized}, where changes can be linked to disaster response scenarios (e.g. damage detection) \cite{gueguen2015large} or monitoring of land cover dynamics \cite{khan2017forest,vaduva2013latent}. Prior approaches often rely on unsupervised methods for change detection, e.g. image differencing, due to high cost of obtaining ground-truth annotations \cite{bruzzone2000automatic}. 
Notably, \cite{gueguen2015large} propose a semi-supervised approach with human in the loop, relying on a hierarchical shape representation.

Another prominent domain is street scenes  \cite{alcantarilla2018street,kataoka2016semantic}. Notably, \cite{sakurada2015change} propose a Panoramic Change Detection Dataset, built off Google Street View panoramic images. In their follow-up work,  \cite{sakurada2017dense} propose an approach to change detection which relies on dense optical flow to address the difference in viewpoints between the images. In a recent work, \cite{palazzolo2018fast} rely on 3D models to identify scene changes by re-projecting images on one another. 
Another line of work targets change detection in video, e.g. using a popular CDnet benchnmark \cite{goyette2012changedetection,wang2014cdnet}, where background subtraction is a successful strategy \cite{bianco2017far}. 
Instead of relying on costly pixel-level video annotation, \cite{khan2017ijcai} propose a weakly supervised approach, which estimates pixel-level labels with a CRF.

Other works address a more subtle, fine-grained change detection, where an object may change its appearance over time, e.g. for the purpose of a valuable object monitoring \cite{feng2015fine,huang2017learning}. To tackle this problem, \cite{stent2016precise} estimate a dense flow field between images to address viewpoint differences. 

Our \MODELABBR{} model relies on an attention mechanism rather than pixel-level difference or flow. Besides, our task is not only to detect the changes, but also to describe them in natural language, going beyond the discussed prior works.

\myparagraph{Task-specific caption generation}
While most image captioning works focus on a generic task of obtaining image relevant descriptions \cite{anderson2018bottom,donahue15cvpr,vinyals2015show}, some recent works explore pragmatic or ``task-specific'' captions. Some focus on generating textual explanations for deep models' predictions \cite{hendricks2016generating,hendricks2018grounding,park2018multimodal}. Others aim to generate a discriminative caption for an image or image region, to disambiguate it from a distractor \cite{andreas2016reasoning,cohn2018pragmatically,luo2017comprehension,luo2018discriminability,vedantam2017context,yu2017joint}. This is relevant to our work, as part of the change caption serves as a referring expression to put an object in context of the other objects. However, our primary focus is to correctly describe the scene changes.

The most related to ours is the work of \cite{jhamtani2018learning}, who also address the task of change captioning for a pair of images. %
While we aim to distinguish distractors from relevant changes, they assume there is always a change between the two images. Next, their pixel-difference based approach assumes that the images are aligned, while we tackle viewpoint change between images. Finally, we systematically study different change types in our new \DATASETNAME{}. We show that our approach generalizes to their Spot-the-Diff dataset in \autoref{sec:spot-the-diff}.

\begin{figure*}[t]
\vspace{-0.4cm}
\includegraphics[width=\linewidth]{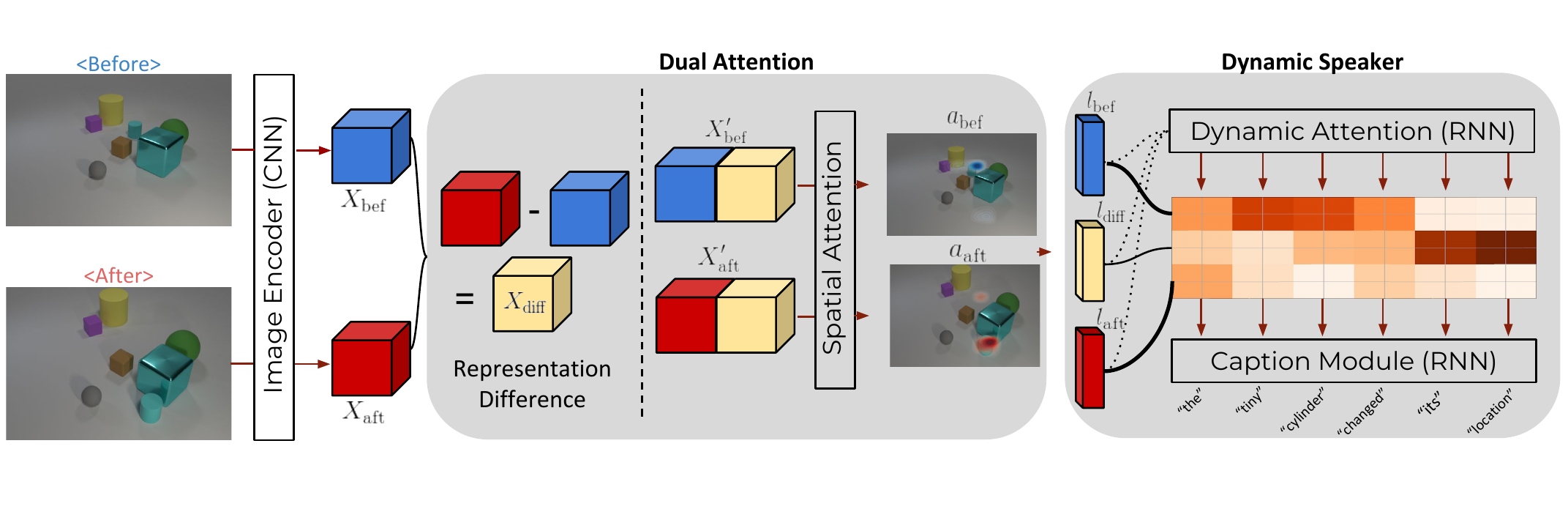}
\vspace{-1.0cm}
\caption{Our \MODELNAME{} (\MODELABBR{}) consists of two main components: Dual Attention (\autoref{subsec:localizer}) and Dynamic Speaker (\autoref{subsec:speaker}).}
\label{fig:model}
\end{figure*}

\myparagraph{Attention in image captioning}
Attention mechanism \cite{bahdanau2015neural} over the visual features  was first used for image captioning by \cite{xu2015show}. Multiple works have since adopted and extended this approach \cite{fu2017aligning,lu2017knowing,pedersoli2016areas}, including performing attention over object detections \cite{anderson2018bottom}. 
Our \MODELABBR{} model relies on two forms of attention: \emph{spatial} Dual Attention used to localize changes between two images, and \emph{semantic} attention, used by our Dynamic Speaker to adaptively focus on  ``before'', ``after'' or ``difference'' visual representations.

\section{\label{sec:approach} \MODELNAME{} (\MODELABBR{})}
We propose a \emph{\MODELNAME{} (\MODELABBR{})} for change detection and captioning. Given a pair of ``before'' and ``after'' images ($I_{\text{bef}}$ and $I_{\text{aft}}$, respectively), our model first detects whether a scene change has happened, and if so, locates the change on both $I_{\text{bef}}$ and $I_{\text{aft}}$. The model then generates a sentence that not only correctly describes the change, but also is spatially and temporally grounded in the image pair. 
To this end, our model includes a Dual Attention (localization) component, followed by a Dynamic Speaker component to generate change descriptions. 
An overview of our model is shown in \autoref{fig:model}.

We describe the implementation details of our Dual Attention in \autoref{subsec:localizer}, and our Dynamic Speaker in \autoref{subsec:speaker}. In \autoref{subsec:training}, we detail our training procedure for jointly optimizing both components using change captions as the only supervision.

\vspace{-0.5cm}
\subsection{\label{subsec:localizer} Dual Attention}
Our Dual Attention acts as a change localizer between $I_{\text{bef}}$ and $I_{\text{aft}}$. Formally, it is a function $f_{\text{loc}}(X_{\text{bef}}, X_{\text{aft}}; \theta_{\text{loc}})=(l_{\text{bef}}, l_{\text{aft}})$ parameterized by $\theta_{\text{loc}}$ that takes $X_{\text{bef}}$ and $X_{\text{aft}}$ as inputs, and outputs feature representations $l_{\text{bef}}$ and $l_{\text{aft}}$ that encode the change manifested in the input pairs. In our implementation, $X_{\text{bef}}, X_{\text{aft}} \in \mathbb{R}^{C \times H \times W}$ are image features of $I_{\text{bef}}, I_{\text{aft}}$, respectively, encoded by a pretrained ResNet~\cite{he2016deep}. 

We first subtract $X_{\text{bef}}$ from $X_{\text{aft}}$ in order to capture semantic difference in the representation space. The resulting tensor $X_\text{diff}$ is concatenated with both $X_{\text{bef}}$ and $X_{\text{aft}}$ which are then used to generate two separate spatial attention maps $a_\text{bef}, a_\text{aft} \in \mathbb{R}^{1 \times H \times W}$. Following \cite{mascharka2018transparency}, we utilize elementwise \textit{sigmoid} instead of \textit{softmax} for computing our attention maps to avoid introducing any form of global normalization. Finally, $a_\text{bef}$ and $a_\text{aft}$ are applied to the input features to do a weighted-sum pooling over the spatial dimensions:
\begin{gather}
    X_\text{diff} = X_\text{aft} - X_\text{bef} \\
    X^\prime_\text{bef} = [X_\text{bef}\;;\;X_\text{diff}], X^\prime_\text{aft} = [X_\text{aft}\;;\;X_\text{diff}] \\
    a_\text{bef} = \sigma(\text{conv}_2(\text{ReLU}(\text{conv}_1(X^\prime_\text{bef})))) \\
    a_\text{aft} = \sigma(\text{conv}_2(\text{ReLU}(\text{conv}_1(X^\prime_\text{aft})))) \\
    l_{\text{bef}} = \sum_{H,W} a_\text{bef} \odot X_\text{bef}, \, l_{\text{bef}} \in \mathbb{R}^{C}\\
    l_{\text{aft}} = \sum_{H,W} a_\text{aft} \odot X_\text{aft}, \, l_{\text{aft}} \in \mathbb{R}^{C}
\end{gather}
where $[;]$, $\text{conv}$, $\sigma$, and $\odot$ indicate concatenation, convolutional layer, elementwise \textit{sigmoid}, and elementwise multiplication, respectively. 
See \autoref{fig:model} for the visualization of Dual Attention component.

This particular architectural design allows the system to attend to images differently depending on the type of a change and the amount of a viewpoint shift, which is a capability crucial for our task. For instance, to correctly describe that an object has moved, the model needs to localize and match the moved object in \textit{both} images; having single attention that locates the object only in one of the images is likely to cause confusion between e.g. moving vs. adding an object. Even if there is an attribute change (e.g. color) which does not involve object displacement, single attention might not be enough to correctly localize the changed object under a viewpoint shift. Unlike ~\cite{yang2016stacked, nam2016dual, lu2016hierarchical, kim2017interpretable, park2018multimodal}, \MODELABBR{} utilizes Dual Attention to process \textit{multiple} visual inputs separately and thereby addresses \TASKNAME{} in the presence of distractors.

\subsection{\label{subsec:speaker} Dynamic Speaker}
Our Dynamic Speaker is based on the following intuition: in order to successfully describe a change, the model should not only learn \textit{where} to look in each image (\emph{spatial} attention, predicted by the Dual Attention), but also \textit{when} to look at each image (\emph{semantic} attention, here). Ideally, we would like the model to exhibit dynamic reasoning, where it learns when to focus on ``before'' ($l_{\text{bef}}$), ``after'' ($l_{\text{aft}}$), or ``difference'' feature ($l_{\text{diff}} = l_{\text{aft}} - l_{\text{bef}}$) as it generates a sequence of words. For example, it is necessary to look at the ``after'' feature ($l_{\text{aft}}$) when referring to a new object added to a scene. \autoref{fig:model} illustrates this behaviour.

To this end, our Dynamic Speaker predicts an attention $\alpha_{i}^{(t)}$ over the visual features $l_{i}$'s at each time step $t$, and obtains the dynamically attended feature $l_\text{dyn}^{(t)}$:
\begin{equation} \label{eq:7}
l_\text{dyn}^{(t)} = \sum_{i}\alpha_{i}^{(t)} l_{i}
\end{equation}
where $i \in (\text{bef}, \text{diff}, \text{aft})$. We use the attentional Recurrent Neural Network~\cite{bahdanau2014neural} to model this formulation. 

Our Dynamic Speaker consists of two modules, namely the dynamic attention module and the caption module. Both are recurrent models based on LSTM~\cite{hochreiter1997long}. At each time step $t$, the LSTM decoder in the dynamic attention module takes as input the previous hidden state of the caption module $h_{c}^{(t-1)}$ and some latent projection $v$ of the visual features $l_{\text{bef}}$, $l_{\text{diff}}$, and $l_{\text{aft}}$ to predict attention weights $\alpha_{i}^{(t)}$:
\begin{equation}
v = \text{ReLU}(W_{d_1}[l_{\text{bef}} \;;\; l_{\text{diff}} \;;\; l_{\text{aft}}] + b_{d_1})
\end{equation}
\begin{equation}
u^{(t)} = [v \;;\; h_{c}^{(t-1)}]
\end{equation} 
\begin{equation}
h_{d}^{(t)} = \text{LSTM}_{d}(h_{d}^{(t)} | u^{(t)}, h_{d}^{(0:t-1)})
\end{equation}
\begin{equation}\label{eq:11}
\alpha^{(t)} \sim \text{Softmax}(W_{d_2} h_{d}^{(t)} + b_{d_2})
\end{equation}
where $h_d^{(t)}$ and $h_c^{(t)}$ are LSTM outputs at decoder time step $t$ for dynamic attention module and caption module, respectively, and $W_{d_1}$, $b_{d_1}$, $W_{d_2}$, and $b_{d_2}$ are learnable parameters. Using the attention weights predicted from Equation \eqref{eq:11}, the dynamically attended feature  $l_\text{dyn}^{(t)}$ is obtained according to Equation \eqref{eq:7}. Finally, $l_\text{dyn}^{(t)}$ and the embedding of the previous word $w_{t-1}$ (ground-truth word during training, predicted word during inference) are input to the LSTM decoder of the caption module to begin generating distributions over the next word: 
\begin{equation}
    x^{(t-1)} = E \mathbbm{1}_{w_{t-1}}
\end{equation}
\begin{equation}
    c^{(t)} = [x^{(t-1)} \;;\; l_\text{dyn}^{(t)}]
\end{equation}
\begin{equation}
    h_{c}^{(t)} = \text{LSTM}_{c}(h_{c}^{(t)} | c^{(t)}, h_{c}^{(0:t-1)})
\end{equation}
\begin{equation}\label{eq:15}
w_{t} \sim \text{Softmax}(W_{c} h_{c}^{(t)} + b_{c})
\end{equation}
where $\mathbbm{1}_{w_{t-1}}$ is a one-hot encoding of the word $w_{t-1}$, $E$ is an embedding layer, and $W_{c}$, $b_{c}$ are learned parameters.

\subsection{\label{subsec:training} Joint Training}
We jointly train the Dual Attention and the Dynamic Speaker end-to-end by maximizing the likelihood of the observed word sequence. Let $\theta$ denote all the parameters in \MODELABBR{}. For a target ground-truth sequence ($w_1^*, \dots , w_T^*$), the objective is to minimize the cross entropy loss:
\begin{equation}
    L_{XE}(\theta) = - \sum_{t=1}^{T} \log (p_{\theta}(w_t^* | w_1^*, \dots, w_{t-1}^*))
\end{equation}
where $p_{\theta}(w_t | w_1, \dots, w_{t-1})$ is given by Equation \eqref{eq:15}. Similar to \cite{mascharka2018transparency}, we apply $L_1$ regularization to the spatial attention masks generated by our Dual Attention in order to minimize unnecessary activations. We also use an entropy regularization over the attention weights generated by our Dynamic Speaker to encourage exploration in using visual features. The final loss function we optimize is as follows:
\begin{equation}
L(\theta) = L_{XE} + \lambda_{L_{1}} L_{1} - \lambda_{ent} L_{ent}
\end{equation}
where $L_{1}$ and $L_{ent}$ are $L_1$ and entropy regularization, respectively, and $\lambda_{L_{1}}$ and $\lambda_{ent}$ are hyperparameters. 
Note, that the Dual Attention component receives no direct supervision for change localization. The only available supervision is obtained through the Dynamic Speaker, which then directs the Dual Attention towards discovering the change.

\section{\label{sec:dataset} \DATASETNAME{}}

\newcommand{\midruleStats}{\cmidrule(rr){1-1}\cmidrule(rr){2-7} \cmidrule(rr){8-8}}
\begin {table}[t]
\footnotesize
\begin{center}
\begin{tabular}{@{}l@{\ }|@{}r@{\ \ }r@{\ \ }r@{\ \ }r@{\ \ }r@{\ \ }r@{\ \ }r@{}}
\toprule
& {DI} & {C} & {T}  & {A}  & {D}  & {M} & {All}\\
\midruleStats
\# Img Pairs & $39,803$ & $79,58$ & $7,963$ & $7,966$ & $79,61$ & $79,55$ & $79,606$ \\
\# Captions & $199,015$ & $58,850$ & $58,946$ & $59,198$ & $58,843$ & $588,83$ & $493,735$ \\
\# Bboxes & - & $15,916$ & $15,926$ & $7,966$ & $7,961$ & $15,910$ & $64,679$ \\
\bottomrule
\end{tabular}
\end{center}
\vspace{-0.4cm}
\caption {\DATASETABBR{} Dataset statistics: number of image pairs, captions, and bounding boxes for each change type: DISTRACTOR (DI), COLOR (C), TEXTURE (T), ADD (A), DROP (D), MOVE (M).} \label{tab:stats} 
\end {table}

Given a lack of an appropriate dataset to study \TASKNAME{} in the presence of distractors, we build the \DATASETNAME{}, based on the CLEVR engine~\cite{johnson2017clevr}. We choose CLEVR, inspired by many works that use it to build diagnostic datasets for various vision and language tasks, e.g. visual question answering~\cite{johnson2017clevr}, referring expression comprehension~\cite{hu2018explainable,liu2019clevr}, text-to-image generation~\cite{el2018keep} or visual dialog~\cite{kottur2019clevr}. As \TASKNAME{} is an emerging task we believe our dataset can complement existing datasets, e.g. \cite{jhamtani2018learning}, which is small, always assumes the presence of a change and lacks localization ground-truth.

First, we generate random scenes with multiple objects in them, which serve as ``before'' images. Note, that in domains such as satellite
imagery \cite{liu2018change,tian2014building,zanetti2016generalized} or surveillance/street scenes \cite{alcantarilla2018street,kataoka2016semantic,palazzolo2018fast}, typical distractors include changes in camera position/zoom or illumination. Motivated by these applications we approach distractor construction accordingly. For each ``before'' image we create two ``after'' images. In the first one, we change the camera position leading to a different angle, zoom, and/or illumination. We have a specific allowed range for the transformation parameters: for each $(x, y, z)$ camera location, we randomly sample a number from the range between $-2.0$ and $2.0$, and jitter the original coordinates by the sampled amount. In the second ``after'' image, we additionally introduce a \emph{scene change}. We consider the following types of scene changes: (a) an object's color is changed, (b) an object's texture is changed, (c) a new object is added, (d) an existing object is dropped, (e) an existing object is moved. In the following we refer to these as: COLOR, TEXTURE, ADD, DROP, MOVE, and DISTRACTOR for no scene change. In total, we generate $39,803$ ``before'' images with respectively $79,606$ ``after'' images. 
We make sure that the number of data points for each scene change type is balanced. 
The dataset is split into $67,660$, $3,976$, and $7,970$ training/validation/test image pairs, respectively.

\begin{figure}[t]
    \centering
    \vspace{-0.2cm}
    \includegraphics[width=\linewidth]{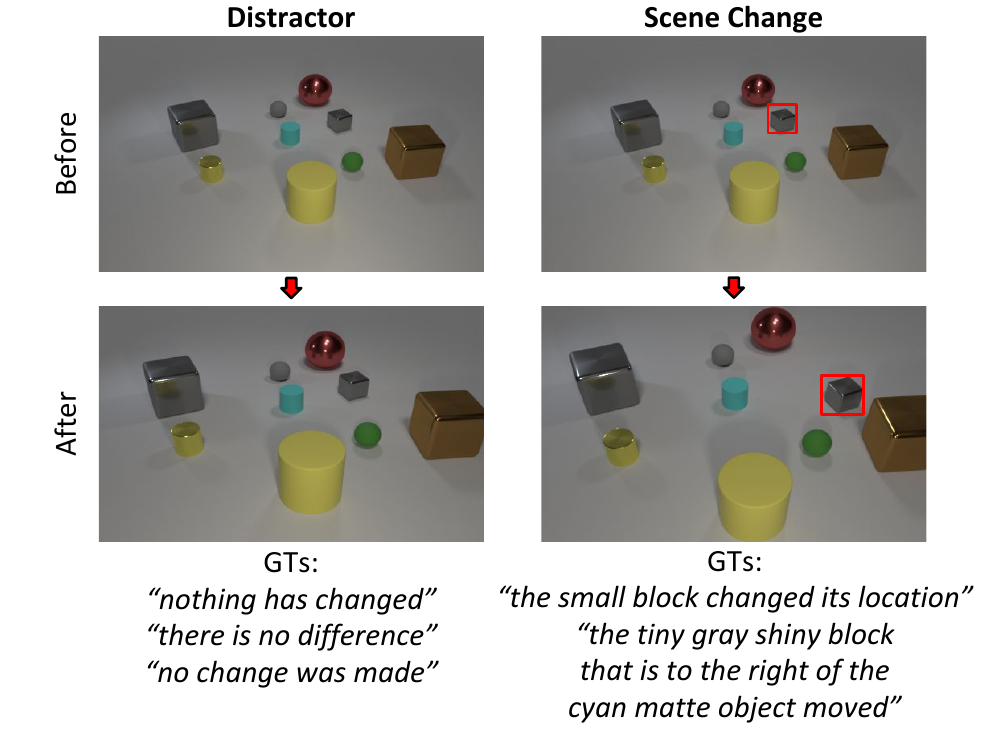}
    \caption{\DATASETABBR{} examples: distractors vs. scene changes, ground-truth captions and bounding boxes.}
    \vspace{-0.2cm}
    \label{fig:dataset_examples}
\end{figure}

In addition to generating the ``before'' and ``after'' scenes, we generate natural language change captions. Each caption is automatically constructed from two parts: the \emph{referring} part (e.g. ``A large blue sphere to the left of a red object'') and the \emph{change} part (e.g. ``has appeared''). Note that for all the change types except ADD, the referring expression is generated based on the ``before'' image, while for ADD, the ``after'' image is used. 
To get the change part, we construct a set of change specific templates (e.g. ``X has been added'', ``X is no longer there'', ``no change was made'' see supplemental for details).%

Finally, we obtain spatial locations of where each scene change took place, so that we can evaluate the correctness of change localization. Specifically, we obtain bounding boxes for all the objects affected by a change, either in one image or in both (``before''/``after''), depending on the change type. 
The overall dataset statistics are shown in \autoref{tab:stats}, and some examples of distractors vs. scene changes with their  descriptions and bounding boxes are shown in \autoref{fig:dataset_examples}.
\section{\label{sec:experiments} Experiments}

{
\setlength{\tabcolsep}{3pt}
\renewcommand{\arraystretch}{1.2}
\begin{table*}[tb]
\begin{center}
\small
\begin{tabular}{l|cccc|cccc|cccc}
\toprule
&\multicolumn{4}{c|}{Total} & \multicolumn{4}{c|}{Scene Change} & \multicolumn{4}{c}{Distractor} \\
Approach & \multicolumn{1}{c}{B} & \multicolumn{1}{c}{C} & \multicolumn{1}{c}{M} & \multicolumn{1}{c|}{S} & \multicolumn{1}{c}{B} & \multicolumn{1}{c}{C} & \multicolumn{1}{c}{M} & \multicolumn{1}{c|}{S} & \multicolumn{1}{c}{B} & \multicolumn{1}{c}{C} & \multicolumn{1}{c}{M} & \multicolumn{1}{c}{S} \\
\midrule
Capt-Pix-Diff & 30.2&	75.9&	23.7&	17.1&	21.9&	36.2&	17.7&	7.9&	43.4&	98.2&	38.9&	26.3 \\
Capt-Rep-Diff & 33.5&	87.9&	26.7&	19.0&	26.0&	51.8&	21.1&	10.1&	49.4&	105.3&	41.7&	27.8 \\
Capt-Att & 42.7&	106.4&	32.1&	23.2&	38.3&	87.2&	27.9&	18.0&	53.5&	106.6&	43.2&	28.4 \\
Capt-Dual-Att & 43.5&	108.5&	32.7&	23.4&	38.5&	89.8&	28.5&	18.2&	56.3&	108.9&	44.0&	28.7 \\
\MODELABBR{} (Ours) & \textbf{47.3}&	\textbf{112.3}&	\textbf{33.9}&	\textbf{24.5}&	\textbf{42.9}&	\textbf{94.6}&	\textbf{29.7}&	\textbf{19.9}&	\textbf{59.8}&	\textbf{110.8}&	\textbf{45.2}&	\textbf{29.1} \\
\bottomrule
\end{tabular}
\caption{\TASKNAME{} evaluation on our \DATASETNAME{}. Our proposed model outperforms all baselines on BLEU-4 (B), CIDEr (C), METEOR (M), and SPICE (S) in each setting (i.e. Total, Scene Change, Distractor).}%
\label{tbl:captioning}
\end{center}
\end{table*}
}

{
\setlength{\tabcolsep}{3pt}
\renewcommand{\arraystretch}{1.2}
\begin{table*}[tb]
\begin{center}
\small
\vspace{-0.4cm}
\begin{tabular}{l|cccccc|cccccc|cccccc}
\toprule
& \multicolumn{6}{c|}{CIDEr} & \multicolumn{6}{c|}{METEOR} & \multicolumn{6}{c}{SPICE} \\
Approach & \multicolumn{1}{c}{C} & \multicolumn{1}{c}{T} & \multicolumn{1}{c}{A} & \multicolumn{1}{c}{D} & \multicolumn{1}{c}{M} &  \multicolumn{1}{c|}{DI} & \multicolumn{1}{c}{C} & \multicolumn{1}{c}{T} & \multicolumn{1}{c}{A} & \multicolumn{1}{c}{D} & \multicolumn{1}{c}{M} &\multicolumn{1}{c|}{DI}& \multicolumn{1}{c}{C} & \multicolumn{1}{c}{T} & \multicolumn{1}{c}{A} & \multicolumn{1}{c}{D} & \multicolumn{1}{c}{M} &\multicolumn{1}{c}{DI} \\
\midrule
Capt-Pix-Diff & 4.2& 16.1&	30.1&	27.1&	18.0& 98.2 & 7.4&	16.0&	24.4&	20.9&	18.2& 38.9 & 1.3&	6.8&	11.4&	10.6&	9.2& 26.3 \\
Capt-Rep-Diff &  44.5&	21.9&	50.1&	49.7&	26.5&	105.3& 19.2&	18.2&	25.7&	23.5&	18.9& 41.7&	8.2&	8.8&	12.1&	12.0&	9.6 & 27.8 \\
Capt-Att & 112.1&	75.9&	91.5&	98.4&	49.6& 106.6&	30.5&	25.4&	30.2&	31.2&	22.2&	43.2& 17.9&	16.3&	19.0&	\textbf{22.3}&	14.5 & 28.4 \\
Capt-Dual-Att & 115.8&	82.7&	85.7&	103.0&	52.6& 108.9&	32.1&	26.7&	29.5&	\textbf{31.7}&	22.4& 44.0&	19.8&	17.6&	16.9&	21.9&	14.7& 28.7 \\
\MODELABBR{} (Ours) & \textbf{120.4}&	\textbf{86.7}&	\textbf{108.2}&	\textbf{103.4}&	\textbf{56.4}& \textbf{110.8}&	\textbf{32.8}&	\textbf{27.3}&	\textbf{33.4}&	31.4&	\textbf{23.5}& \textbf{45.2}&	\textbf{21.2}&	\textbf{18.3}&	\textbf{22.4}&	22.2&	\textbf{15.4}& \textbf{29.1}  \\
\bottomrule
\end{tabular}
\caption{A Detailed breakdown of \TASKNAME{} evaluation on our \DATASETNAME{} by change types: Color (C), Texture (T), Add (A), Drop (D), Move (M), and Distractor (DI).}%
\vspace{-0.6cm}
\label{tbl:captioning_by_type}
\end{center}
\end{table*}
}

In this section, we evaluate our \MODELABBR{} model on the \TASKNAME{} task against a number of baselines. First, we present quantitative results for the ablations and discuss their implications on our new \DATASETNAME{}. We also provide qualitative analysis of the generated captions, examine attention weights predicted by \MODELABBR{}, and assess its robustness to viewpoint shift. Finally, we test the general effectiveness of our approach on the Spot-the-Diff \cite{jhamtani2018learning}, a realistic dataset with no distractors.

\subsection{Experimental setup}
\label{sec:experimentalsetup}
Here, we detail our experimental setup in terms of implementation and evaluation schemes.

\myparagraph{Implementation Details.}
Similar to \cite{DBLP:journals/corr/HuARDS17, DBLP:journals/corr/JohnsonHMHLZG17, DBLP:journals/corr/abs-1709-07871}, we use ResNet-101 \cite{he2016deep} pretrained on ImageNet \cite{deng2009imagenet} to extract visual features from the images. We use features from the convolutional layer right before the global average pooling, obtaining features with dimensionality of $1024$ x $14$ x $14$. The LSTMs used in the Dynamic Speaker have a hidden state dimension of $512$. The word embedding layer is trained from scratch and each word is represented by a $300$-dim vector. We train our model for $40$ epochs using the Adam Optimizer~\cite{kingma2014adam} with a learning rate of $0.001$ and a batch size of $128$. The hyperparameters for the regularization terms are  $\lambda_{L_{1}} = 2.5e^{-03}$ and $\lambda_{ent } = 0.0001$. Our model is implemented using PyTorch~\cite{paszke2017automatic}, and our code and dataset will be made publicly available. 

\myparagraph{Evaluation.}
To evaluate change captioning, we rely on BLEU-4~\cite{papineni2002bleu}, METEOR~\cite{banerjee2005meteor}, CIDEr~\cite{vedantam2015cider}, and SPICE~\cite{anderson2016spice} metrics which measure overall sentence fluency and similarity to ground-truth. For change localization, we rely on the Pointing Game evaluation~\cite{zhang2018top}. We use bilinear interpolation to upsample the attention maps to the original image size, and check whether the point with the highest activation ``falls'' in the ground-truth bounding box.

\subsection{Results on \DATASETNAME{}}

\myparagraph{Pixel vs. representation difference}\label{subsec:pixvsrep}
\cite{jhamtani2018learning} utilize pixel difference information when generating change captions under the assumption that the images are aligned. To obtain insights into whether a similar approach can still be effective when a camera position changes, we introduce the following baselines: \textit{Capt-Pix-Diff} is a model that directly utilizes pixel-wise difference in the RGB space between ``before'' and ``after'' images. We use pyramid reduce downsampling on the RGB difference to match the spatial resolution of the ResNet features. The downsampled tensor is concatenated with the ResNet features on which we apply a series of convolutions and max-pooling. The resulting feature, which combines ``before'', ``after'', and ``pixel difference'' information, is input to an LSTM for sentence generation. On the other hand, \textit{Capt-Rep-Diff} relies on representation difference (i.e. $X_{\text{diff}}$) instead of pixel difference. A series of convolutions and max-pooling are applied to the representation difference and then input to an LSTM decoder. As shown in the first two rows of \autoref{tbl:captioning}, \textit{Capt-Rep-Diff} outperforms \textit{Capt-Pix-Diff} in all settings, indicating that representation difference is more informative than pixel difference when comparing scenes under viewpoint shift. We believe this is because visual representations are more semantic by nature, and each activation in the representation has a larger receptive field that allows the difference operation to be less sensitive to the camera shift. As a result, we deliberately use representation difference in all subsequent experiments.

\myparagraph{Role of localization}\label{subsec:localization}
To understand the importance of localization for change description, we compare models with and without spatial attention mechanism. \textit{Capt-Att} is an extension of \textit{Capt-Rep-Diff} which learns a single spatial attention which is applied to both ``after'' and ``before'' features. The attended features are subtracted and input to an LSTM decoder. We observe that \textit{Capt-Att} significantly outperforms \textit{Capt-Rep-Diff}, indicating that the capacity to explicitly localize the change has a high impact on the caption quality in general. Note, that the improvements are more pronounced for scene changes (i.e. C, T, A, D, M) than for distractors (DI), see  ~\autoref{tbl:captioning_by_type}, which is intuitive since the localization ability matters most when there actually is a scene change.  %

\begin{table}[t]
\setlength{\tabcolsep}{2pt}
\renewcommand{\arraystretch}{1.2}
\small
\centering
\begin{tabular}{l|ccccc|c}
\toprule
& C & T & A & D & M & Total \\
\midrule
Capt-Att & 46.68 &	57.90&	22.84&	47.80&	17.57&	39.37 \\
Capt-Dual-Att & 40.97&	46.55&	54.33&	45.67&	19.89&	39.35 \\
\MODELABBR{} (Ours) & \textbf{54.52}&	\textbf{65.75}&	\textbf{48.68}&	\textbf{50.06}&	\textbf{22.77}&	\textbf{48.10} \\
\bottomrule
\end{tabular}
\caption{Pointing game accuracy results. We report per change-type performance (Color (C), Texture (T), Add (A), Drop (D), Move (M))
as well as the total performance. The numbers are in \%.}
\vspace{-0.4cm}
\label{tbl:pointing_by_type}
\end{table}

\begin{figure}[t]
\begin{center}
\vspace{-1.0mm}
\includegraphics[width=\linewidth]{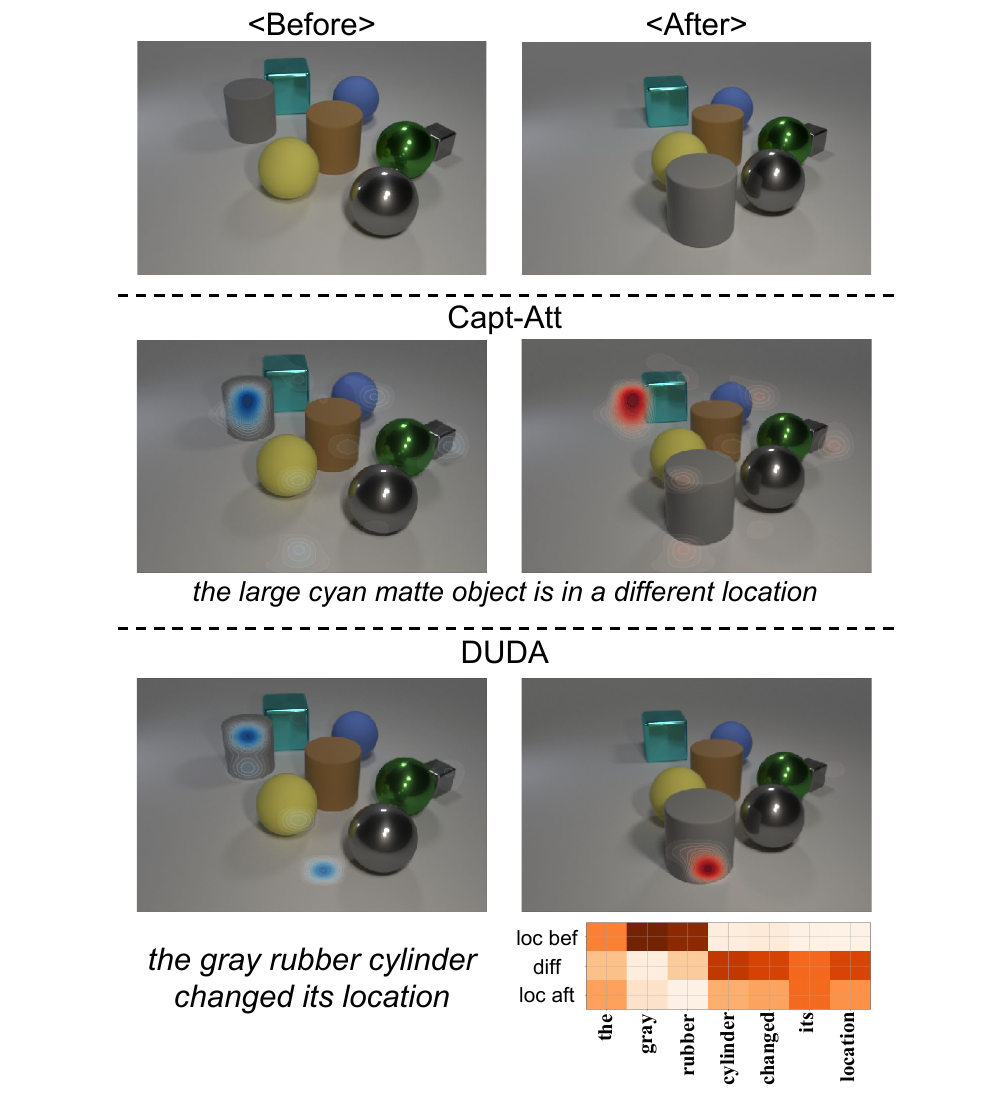}
\end{center}
\vspace{-6.0mm}
\caption{Qualitative results comparing \textit{Capt-Att} and \MODELABBR{}. The blue and red attention maps are applied to ``before'' and ``after'', respectively. The blue and red attention maps are the same for \textit{Capt-Att} whereas in \MODELABBR{} they are separately generated. The heat map on the lower-right is the visualization of the dynamic attention weights where the rows represent the amount of attention given to each visual feature (e.g. loc bef, diff, loc aft) per word.}
\vspace{-0.5cm}
\label{fig:comparison}
\end{figure}

\myparagraph{Single attention vs. dual attention}\label{subsec:singlevsdual}
Using multiple spatial attentions has been shown to be useful for many purposes including multi-step/hierarchical reasoning~\cite{yang2016stacked, nam2016dual, lu2016hierarchical} and model interpretability~\cite{kim2017interpretable, park2018multimodal}. To this extent, we train a model that deploys Dual Attention and evaluate its application to \TASKNAME{} in the presence of distractors.  \textit{Capt-Dual-Att} is an extension of \textit{Capt-Att} which learns two separate spatial attentions for the pair of images. Compared to \textit{Capt-Att}, \textit{Capt-Dual-Att} achieves higher performance overall according to \autoref{tbl:captioning}. However, the improvements are limited in the sense that the margin of increase is small and not all change types improve (see \autoref{tbl:captioning_by_type}). A similar issue can be seen in the Pointing Game results in \autoref{tbl:pointing_by_type}. We speculate that without a proper inductive bias, it is difficult to learn how to utilize two spatial attentions effectively; a more complex speaker that enforces the usage of multiple visual signals might be required. 

\myparagraph{Dual + dynamic attention}\label{subsec:dynamic}
Our final model with the Dynamic Speaker outperforms all previously discussed baselines not only in captioning (\autoref{tbl:captioning}, \autoref{tbl:captioning_by_type}) but also in localization (\autoref{tbl:pointing_by_type}), supporting our intuition above. In \autoref{fig:comparison}, we compare results from \textit{Capt-Att} and \MODELABBR{}. We observe that a single spatial attention used in \textit{Capt-Att} cannot locate and associate the moved object in ``before'' and ``after'' images, thus confusing the properties of the target object (i.e. large \textit{cyan} matte). On the other hand, our model is able to locate and match the target object in both scenes via Dual Attention, and discover that the object has moved. Moreover, it can be seen that our Dynamic Speaker predicts the attention weights that reveal some reasoning capacity of our model, where it first focuses on the ``before'' when addressing the changed object and gradually shifts attention to ``diff'' and ``after'' when mentioning the change. 

\begin{figure}[t]
\begin{center}
\vspace{-1.0mm}
\includegraphics[width=\linewidth]{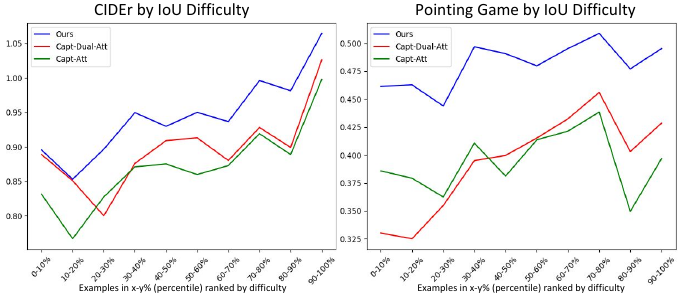}
\end{center}
\vspace{-6.0mm}
\caption{Change captioning and localization performance breakdown by viewpoint shift (measured by IoU).}
\vspace{-0.4cm}
\label{fig:iou}
\end{figure}

\myparagraph{Measuring robustness to viewpoint shift}
\label{subsec:difficult}
The experiments above demonstrate the general effectiveness of our model in tackling the \TASKNAME{} task. We now further validate the robustness of our model to viewpoint shift. To measure the amount of viewpoint shift for a pair of images, we use the following heuristics: for each object in the scene, \textit{excluding} the changed object, we compute the IoU of the object's bounding boxes across the image pair. We assume the more the camera changes its position, the less the bounding boxes will overlap. We compute the mean of these IoUs and sort the test examples based on this (lower IoU means higher difficulty). The performance breakdown in terms of change captioning and localization is shown in \autoref{fig:iou}. Our model outperforms the baselines on both tasks, including the more difficult samples (to the left). We see that both captioning and localization performance degrades for the baselines and our model (although less so) as viewpoint shift increases, indicating that it is an important challenge to be addressed on our dataset. %

\begin{figure*}[t]
\includegraphics[width=\linewidth]{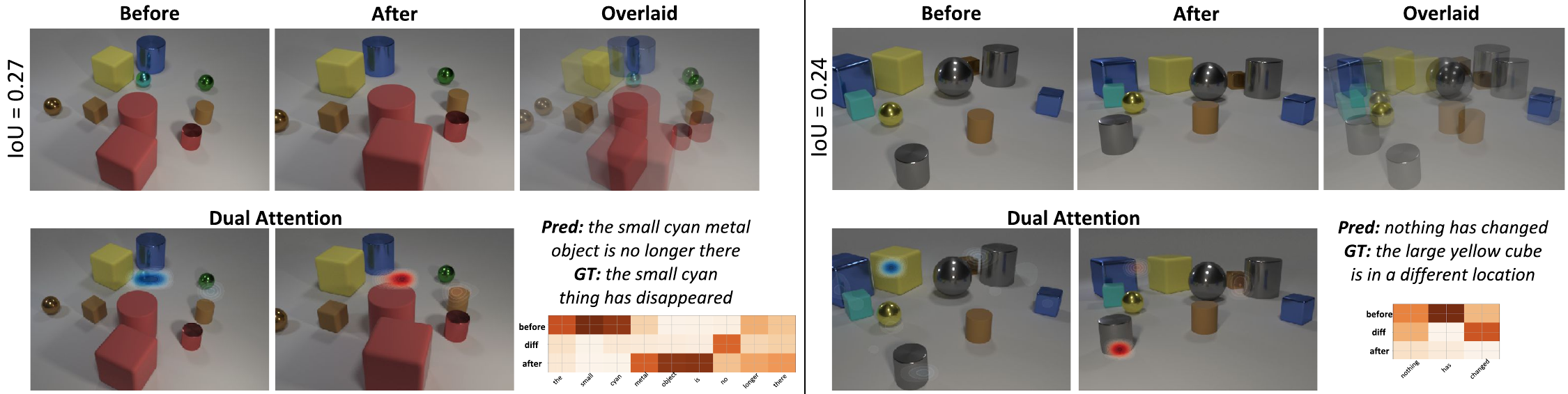}
\vspace{-0.6cm}
\caption{Qualitative examples of \MODELABBR{}. The left is an example in which \MODELABBR{} successfully localizes the change and generates correct descriptions with proper modulations among ``before'', ``diff'', and ``after'' visual features. The right example is a failure case. We observe that significant viewpoint shift leads to incorrect localization of the change, thus confusing the dynamic speaker.}
\label{fig:vidam_results}
\end{figure*}

\autoref{fig:vidam_results} illustrates two examples with large viewpoint changes, as measured by IoU. The overlaid images show that the scale and location of the objects may change significantly. The left example is a success, where  \MODELABBR{} is able to tell that the object has disappeared. Interestingly, in this case, it rarely attends to the ``difference'' feature. The right example illustrates a failure, where \MODELABBR{} predicts that no change has occured, as a viewpoint shift makes it difficult to relate objects between the two scenes. %
Overall, we find that most often the semantic changes are confused with the distractors (no change) rather than among themselves, while MOVE suffers from such confusion the most.

\subsection{Results on Spot-the-Diff Dataset}
\label{sec:spot-the-diff}
We also evaluate our \MODELABBR{} model on the recent Spot-the-Diff dataset \cite{jhamtani2018learning} with real images and human-provided descriptions. This dataset features mostly well aligned image pairs from surveillance cameras, with one or more changes between the images (no distractors). %
We evaluate our model in a single change setting, i.e. we generate a single change description, and use all the available human descriptions as references, as suggested by \cite{jhamtani2018learning}.

\begin{figure}[t]
    \centering
    \includegraphics[width=\linewidth]{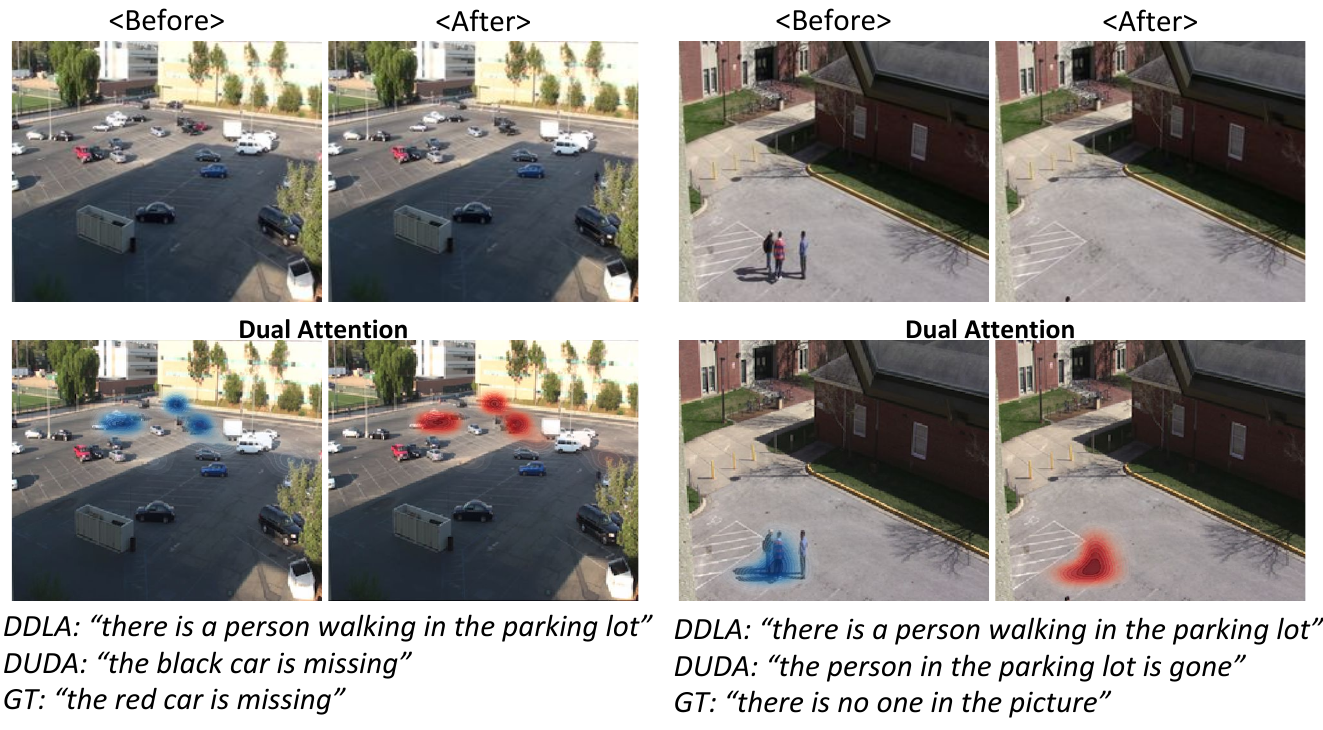}
    \vspace{-0.4cm}
    \caption{Example outputs of our model on the Spot-the-Diff dataset \cite{jhamtani2018learning}. We visualize clusters with the maximum dual attention weights. We also show results from the DDLA~\cite{jhamtani2018learning} and the ground-truth captions.}
    \vspace{-0.3cm}
    \label{fig:spot_the_diff}
\end{figure}

We present our results in \autoref{tbl:EMNLP}. The DDLA approach of \cite{jhamtani2018learning} relies on precomputed spatial clusters, obtained using pixel-wise difference between two images, assuming that the images are aligned. %
For a fair comparison we rely on the same information: we extract visual features from both ``before'' and ``after'' images using the spatial clusters. We apply Dual Attention over the extracted features to learn which clusters should be relayed to the Dynamic Speaker. The rest of our approach is unchanged. As can be seen from \autoref{tbl:EMNLP}, \MODELABBR{} matches or outperforms DDLA on most metrics. We present qualitative comparison in \autoref{fig:spot_the_diff}. As can be seen from the examples, our DUDA model can attend to the right cluster and describe changes corresponding to the localized cluster. 

Despite the usage of natural images and human descriptions, the Spot-the-Diff dataset is not the definitive test for robust change captioning as it does not consider the presence of distractors. That is, one does not have to establish whether the change occurred as there is always a change between each pair of images, and the images are mostly well-aligned. We advocate for a more practical setting of robust change captioning, where determining whether the change is by itself relevant is an important part of the problem. 

{
\begin{table}[tb]
\begin{center}
\small
\begin{tabular}{@{}l|cccc@{}}%
\toprule
Approach & \multicolumn{1}{c}{B} & \multicolumn{1}{c}{C} & \multicolumn{1}{c}{M} & \multicolumn{1}{c}{R} \\
\midrule
DDLA*~\cite{jhamtani2018learning} & 0.081 & 0.340 & 0.115 & 0.283 \\
\MODELABBR{}* & 0.081&	0.325&	0.118&	0.291\\
\bottomrule
\end{tabular}
\caption{
We evaluate our approach on the Spot-the-Diff dataset~\cite{jhamtani2018learning}. * We report results averaged over two runs,for DDLA~\cite{jhamtani2018learning}, we use the two sets of results reported by the authors. See text for details.
}
\vspace{-0.8cm}
\label{tbl:EMNLP}
\end{center}
\end{table}
}

\section{\label{sec:conclusion} Conclusion}

In this work, we address robust \TASKNAME{} in the setting that includes distractors. We propose the novel \MODELNAME to jointly localize and describe changes between images. Our dynamic attention scheme is superior to the baselines and its visualization provides an interpretable view on the change caption generation mechanism. Our model is robust to distractors in the sense that it can distinguish relevant scene changes from illumination/viewpoint changes. Our \DATASETNAME is a new benchmark, where many challenges need to be addressed, e.g. establishing correspondences between the objects in the presence of viewpoint shift, resolving ambiguities and correctly referring to objects in complex scenes, and localizing the changes in the scene amidst viewpoint shifts. Our findings inform us of important challenges in domains like street scenes, e.g. ``linking'' the moved objects in before/after images, as also noted in \cite{jhamtani2018learning}. Our results on Spot-the-Diff are complementary to those we have obtained on the larger CLEVR-Change dataset. While Spot-the-Diff is based on real images, there are minimal or no distractor cases in the dataset. This suggests that valuable future work will be to collect real-image datasets with images undergoing significant semantic and distractor changes. 

\section*{Acknowledgements}
This work was in part supported by the US DoD and the Berkeley Artificial Intelligence Research (BAIR) Lab.
\section*{Supplementary Material}

\appendix
In this supplementary material, we provide an analysis of the performance of our \MODELNAME (\MODELABBR) in terms of what change types get confused the most. We also provide additional details on how \DATASETNAME was collected, especially how change descriptions were generated, and how the data distribution in terms of difficulty measured by IoU looks like given the introduced random jitters in camera position. 

\section{Confusion Matrix of Change Types}
\begin{figure}
    \centering
    \includegraphics[width=0.5\textwidth]{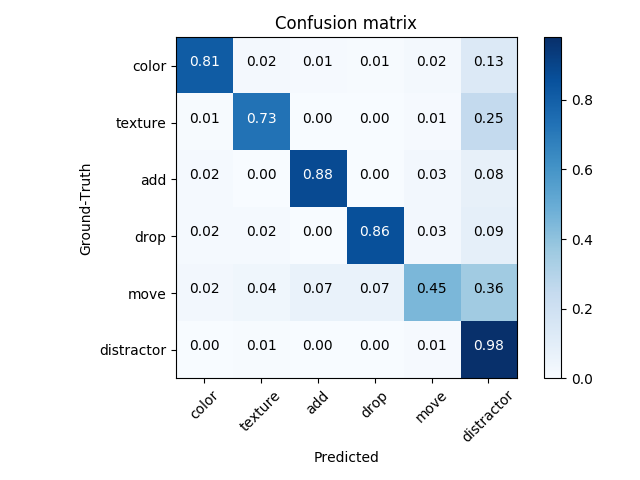}
    \caption{Confusion matrix of \MODELABBR. The horizontal axis indicates the predicted change types of our model whereas the vertical axis indicates the actual change types.}
    \label{fig:confusion}
\end{figure}

In order to analyze the behavior of our method on different change types, we parse the sentences generated by our model and categorize the type of change that is detected based on the parsed results. We compare that to the ground-truth change type information, and plot the confusion matrix in \autoref{fig:confusion}. As we have already shown (\autoref{tbl:captioning_by_type} in the main paper), the most challenging change types are TEXTURE ($73\%$ accuracy) and MOVE ($45\%$ accuracy), which are most often confused with the DISTRACTOR changes. 
It is interesting to note that for all change types most of the confusion comes from misidentifying scene changes as DISTRACTORs, and that such confusion is the most severe for MOVE. This is intuitive in the sense that in order to correctly distinguish MOVE from DISTRACTOR, the model has to spatially relate every other object in the scene whereas for other scene change types the changes are relatively salient and do not necessarily require understanding the spatial relationships between the objects. Moreover, MOVE is also confused with ADD and DROP, as it may be difficult to correctly establish a correspondence between all the objects in ``before'' and ``after'' scenes. Overall, the substantial amount of confusion with the DISTRACTORs demonstrates the difficulty of our problem statement, as opposed to always assuming that a scene change is present.

\section{Additional Details on \DATASETNAME}
In this section, we provide details on how the captions are generated in our \DATASETNAME and how the random camera position shifts manifest themselves in the dataset distribution. Having access to all the object information in a CLEVR-rendered scene, we can easily generate multiple different sentences describing a particular change by using templates listed in \autoref{tab:templates}. For instance once the images are generated with the desired change (e.g. COLOR), we identify the changed object in the before or after images, and extract its locations and attributes which are used to generate a referring expression (e.g. \emph{the red metallic cube that is to the left of a big sphere}). This phrase is then combined with a randomly selected template followed by a description of how it has changed (i.e. \emph{... changed to yellow}). 

\begin {table}[t]
\footnotesize
\begin{center}
\begin{tabular}{@{}l l@{}}
\toprule
Type & Templates \\
\midrule
\multirow{2}{*}{COLOR} & ... changed to ... \\
  & ... turned ... \\
  & ... became ... \\
\midrule
\multirow{2}{*}{TEXTURE}  & ... changed to ... \\
  & ... turned ... \\
  & ... became ... \\
\midrule
\multirow{2}{*}{ADD}  & ... has appeared. \\
  & ... has been newly placed. \\
  & ... has been added. \\
\midrule
\multirow{2}{*}{DROP}  & ... has disappeared. \\
  & ... is missing. \\
  & ... is gone. \\
  & ... is no longer there. \\
\midrule
\multirow{2}{*}{MOVE} & ... moved. \\
  & ... is in a different location. \\
  & ... changed its location. \\
\midrule
\multirow{2}{*}{DISTRACTOR} & no change was made. \\
  & the scene is the same as before. \\
  & the two scenes seem identical. \\
\bottomrule
\end{tabular}
\end{center}
\vspace{-0.4cm}
\caption {For each change type we construct a few templates, based on which the change part of the caption is obtained.} \label{tab:templates} 
\end {table}

\begin{figure*}
    \centering
    \includegraphics[width=1\textwidth]{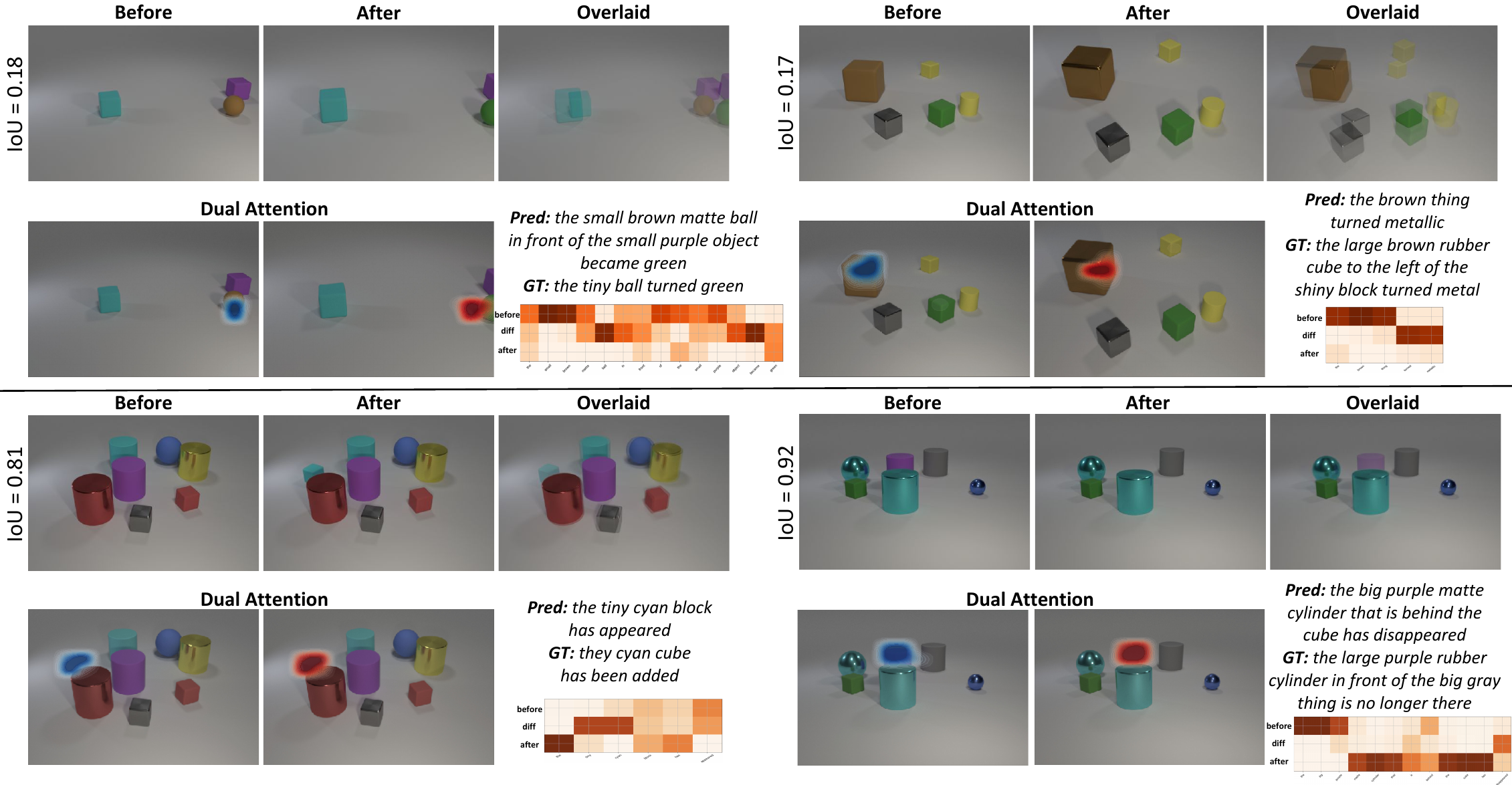}
    \caption{Difficult and easy examples chosen via IoU-based heuristics. The examples at the top are the difficult ones, where the viewpoint shift is noticeable. The examples at the bottom are the easy ones, where the viewpoint change is not significant. We also show the corresponding attention and sentences generated by our model, as well as the ground-truth descriptions.}
    \label{fig:diff_easy_examples}
\end{figure*}

\begin{figure}
    \centering
    \includegraphics[width=0.5\textwidth]{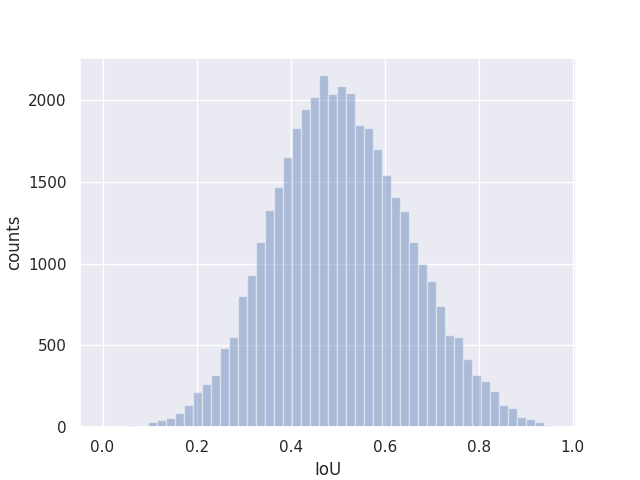}
    \caption{Histogram of \DATASETNAME based on IoU. The horizontal axis indicates the amount of viewpoint shift measured by IoU whereas the vertical axis indicates the number of data points.}
    \label{fig:histogram}
\end{figure}

In \autoref{sec:dataset} of the main paper, it is described that different viewpoint and illumination are introduced via a random shift in camera ($x$, $y$, $z$) location ranging between -2.0 to 2.0 in coordinate points. As a way to understand how this translates to an actual difference between before and after images, we plot a histogram of the entire dataset based on the IoU heuristics explained in the main paper. As can be seen from \autoref{fig:histogram}, the random camera jitters form a reasonable distribution of data points in terms of viewpoint shift difficulty. To better illustrate what the IoU means, we provide relatively difficult (i.e. low IoU of $0.17$ - $0.18$) and easy (i.e. high IoU of $0.81$ - $0.92$) examples in \autoref{fig:diff_easy_examples}. We notice that depending on the viewpoint shift, the task can become significantly difficult even for a simple scene. For instance in the top-left example of \autoref{fig:diff_easy_examples}, where there are only three objects, we see that it becomes hard to localize the changed object as it escapes the scene due to significant camera movement. On the other hand, for a more complex scene like the bottom-left example, localizing change is easier with a small viewpoint shift.

{\small
\bibliographystyle{ieee}
\bibliography{biblioLong,egbib}

\begin{thebibliography}{10}\itemsep=-1pt

\bibitem{alcantarilla2018street}
P.~F. Alcantarilla, S.~Stent, G.~Ros, R.~Arroyo, and R.~Gherardi.
\newblock Street-view change detection with deconvolutional networks.
\newblock {\em Autonomous Robots}, 42(7):1301--1322, 2018.

\bibitem{anderson2016spice}
P.~Anderson, B.~Fernando, M.~Johnson, and S.~Gould.
\newblock Spice: Semantic propositional image caption evaluation.
\newblock In {\em Proceedings of the European Conference on Computer Vision
  (ECCV)}. Springer, 2016.

\bibitem{anderson2018bottom}
P.~Anderson, X.~He, C.~Buehler, D.~Teney, M.~Johnson, S.~Gould, and L.~Zhang.
\newblock Bottom-up and top-down attention for image captioning and visual
  question answering.
\newblock In {\em Proceedings of the IEEE Conference on Computer Vision and
  Pattern Recognition (CVPR)}, 2018.

\bibitem{andreas2016reasoning}
J.~Andreas and D.~Klein.
\newblock Reasoning about pragmatics with neural listeners and speakers.
\newblock In {\em Proceedings of the Conference on Empirical Methods in Natural
  Language Processing (EMNLP)}, 2016.

\bibitem{bahdanau2014neural}
D.~Bahdanau, K.~Cho, and Y.~Bengio.
\newblock Neural machine translation by jointly learning to align and
  translate.
\newblock {\em arXiv:1409.0473}, 2014.

\bibitem{bahdanau2015neural}
D.~Bahdanau, K.~Cho, and Y.~Bengio.
\newblock Neural machine translation by jointly learning to align and
  translate.
\newblock In {\em Proceedings of the International Conference on Learning
  Representations (ICLR)}, 2015.

\bibitem{banerjee2005meteor}
S.~Banerjee and A.~Lavie.
\newblock Meteor: An automatic metric for mt evaluation with improved
  correlation with human judgments.
\newblock In {\em Proceedings of the ACL Workshop on Intrinsic and Extrinsic
  Evaluation Measures for Machine Translation and/or Summarization}, pages
  65--72, 2005.

\bibitem{bianco2017far}
S.~Bianco, G.~Ciocca, and R.~Schettini.
\newblock How far can you get by combining change detection algorithms?
\newblock In {\em International Conference on Image Analysis and Processing},
  pages 96--107. Springer, 2017.

\bibitem{bruzzone2000automatic}
L.~Bruzzone and D.~F. Prieto.
\newblock Automatic analysis of the difference image for unsupervised change
  detection.
\newblock {\em IEEE Transactions on Geoscience and Remote sensing},
  38(3):1171--1182, 2000.

\bibitem{cohn2018pragmatically}
R.~Cohn-Gordon, N.~Goodman, and C.~Potts.
\newblock Pragmatically informative image captioning with character-level
  reference.
\newblock In {\em Proceedings of the Conference of the North American Chapter
  of the Association for Computational Linguistics (NAACL)}, 2018.

\bibitem{deng2009imagenet}
J.~Deng, W.~Dong, R.~Socher, L.-J. Li, K.~Li, and L.~Fei-Fei.
\newblock Imagenet: A large-scale hierarchical image database.
\newblock In {\em Proceedings of the IEEE Conference on Computer Vision and
  Pattern Recognition (CVPR)}. Ieee, 2009.

\bibitem{donahue15cvpr}
J.~Donahue, L.~A. Hendricks, S.~Guadarrama, M.~Rohrbach, S.~Venugopalan,
  K.~Saenko, and T.~Darrell.
\newblock Long-term recurrent convolutional networks for visual recognition and
  description.
\newblock In {\em Proceedings of the IEEE Conference on Computer Vision and
  Pattern Recognition (CVPR)}, 2015.

\bibitem{el2018keep}
A.~El-Nouby, S.~Sharma, H.~Schulz, D.~Hjelm, L.~E. Asri, S.~E. Kahou,
  Y.~Bengio, and G.~W. Taylor.
\newblock Keep drawing it: Iterative language-based image generation and
  editing.
\newblock In {\em Advances in Neural Information Processing Systems Workshops
  (NIPS Workshops)}, 2018.

\bibitem{feng2015fine}
W.~Feng, F.-P. Tian, Q.~Zhang, N.~Zhang, L.~Wan, and J.~Sun.
\newblock Fine-grained change detection of misaligned scenes with varied
  illuminations.
\newblock In {\em Proceedings of the IEEE International Conference on Computer
  Vision (ICCV)}, pages 1260--1268, 2015.

\bibitem{fu2017aligning}
K.~Fu, J.~Jin, R.~Cui, F.~Sha, and C.~Zhang.
\newblock Aligning where to see and what to tell: Image captioning with
  region-based attention and scene-specific contexts.
\newblock {\em IEEE Transactions on Pattern Analysis and Machine Intelligence
  (TPAMI)}, 39(12):2321--2334, 2017.

\bibitem{goyette2012changedetection}
N.~Goyette, P.-M. Jodoin, F.~Porikli, J.~Konrad, P.~Ishwar, et~al.
\newblock Changedetection. net: A new change detection benchmark dataset.
\newblock In {\em Proceedings of the IEEE Conference on Computer Vision and
  Pattern Recognition Workshops (CVPR Workshops)}, 2012.

\bibitem{gueguen2015large}
L.~Gueguen and R.~Hamid.
\newblock Large-scale damage detection using satellite imagery.
\newblock In {\em Proceedings of the IEEE Conference on Computer Vision and
  Pattern Recognition (CVPR)}, pages 1321--1328, 2015.

\bibitem{he2016deep}
K.~He, X.~Zhang, S.~Ren, and J.~Sun.
\newblock Deep residual learning for image recognition.
\newblock In {\em Proceedings of the IEEE Conference on Computer Vision and
  Pattern Recognition (CVPR)}, 2016.

\bibitem{hendricks2016generating}
L.~A. Hendricks, Z.~Akata, M.~Rohrbach, J.~Donahue, B.~Schiele, and T.~Darrell.
\newblock Generating visual explanations.
\newblock In {\em Proceedings of the European Conference on Computer Vision
  (ECCV)}. Springer, 2016.

\bibitem{hendricks2018grounding}
L.~A. Hendricks, R.~Hu, T.~Darrell, and Z.~Akata.
\newblock Grounding visual explanations.
\newblock In {\em Proceedings of the European Conference on Computer Vision
  (ECCV)}, 2018.

\bibitem{hochreiter1997long}
S.~Hochreiter and J.~Schmidhuber.
\newblock Long short-term memory.
\newblock {\em Neural computation}, 9(8):1735--1780, 1997.

\bibitem{hu2018explainable}
R.~Hu, J.~Andreas, T.~Darrell, and K.~Saenko.
\newblock Explainable neural computation via stack neural module networks.
\newblock In {\em Proceedings of the European Conference on Computer Vision
  (ECCV)}, pages 53--69, 2018.

\bibitem{DBLP:journals/corr/HuARDS17}
R.~Hu, J.~Andreas, M.~Rohrbach, T.~Darrell, and K.~Saenko.
\newblock Learning to reason: End-to-end module networks for visual question
  answering.
\newblock In {\em Proceedings of the IEEE International Conference on Computer
  Vision (ICCV)}, 2017.

\bibitem{huang2017learning}
R.~Huang, W.~Feng, Z.~Wang, M.~Fan, L.~Wan, and J.~Sun.
\newblock Learning to detect fine-grained change under variant imaging
  conditions.
\newblock In {\em Proceedings of the IEEE International Conference on Computer
  Vision Workshops (ICCV Workshops)}, 2017.

\bibitem{jhamtani2018learning}
H.~Jhamtani and T.~Berg-Kirkpatrick.
\newblock Learning to describe differences between pairs of similar images.
\newblock In {\em Proceedings of the Conference on Empirical Methods in Natural
  Language Processing (EMNLP)}, 2018.

\bibitem{johnson2017clevr}
J.~Johnson, B.~Hariharan, L.~van~der Maaten, L.~Fei-Fei, C.~L. Zitnick, and
  R.~Girshick.
\newblock Clevr: A diagnostic dataset for compositional language and elementary
  visual reasoning.
\newblock In {\em Proceedings of the IEEE Conference on Computer Vision and
  Pattern Recognition (CVPR)}, 2017.

\bibitem{DBLP:journals/corr/JohnsonHMHLZG17}
J.~Johnson, B.~Hariharan, L.~van~der Maaten, J.~Hoffman, F.~Li, C.~L. Zitnick,
  and R.~B. Girshick.
\newblock Inferring and executing programs for visual reasoning.
\newblock In {\em Proceedings of the IEEE International Conference on Computer
  Vision (ICCV)}, 2017.

\bibitem{kataoka2016semantic}
H.~Kataoka, S.~Shirakabe, Y.~Miyashita, A.~Nakamura, K.~Iwata, and Y.~Satoh.
\newblock Semantic change detection with hypermaps.
\newblock {\em arXiv:1604.07513}, 2016.

\bibitem{khan2017forest}
S.~H. Khan, X.~He, F.~Porikli, and M.~Bennamoun.
\newblock Forest change detection in incomplete satellite images with deep
  neural networks.
\newblock {\em IEEE Transactions on Geoscience and Remote Sensing},
  55(9):5407--5423, 2017.

\bibitem{khan2017ijcai}
S.~H. Khan, X.~He, F.~Porikli, M.~Bennamoun, F.~Sohel, and R.~Togneri.
\newblock Learning deep structured network for weakly supervised change
  detection.
\newblock In {\em Proceedings of the International Joint Conference on
  Artificial Intelligence (IJCAI)}, 2017.

\bibitem{kim2017interpretable}
J.~Kim and J.~F. Canny.
\newblock Interpretable learning for self-driving cars by visualizing causal
  attention.
\newblock In {\em Proceedings of the IEEE International Conference on Computer
  Vision (ICCV)}, 2017.

\bibitem{kingma2014adam}
D.~P. Kingma and J.~Ba.
\newblock Adam: A method for stochastic optimization.
\newblock {\em arXiv:1412.6980}, 2014.

\bibitem{kottur2019clevr}
S.~Kottur, J.~M. Moura, D.~Parikh, D.~Batra, and M.~Rohrbach.
\newblock Clevr-dialog: A diagnostic dataset for multi-round reasoning in
  visual dialog.
\newblock In {\em Proceedings of the Conference of the North American Chapter
  of the Association for Computational Linguistics (NAACL)}, 2019.

\bibitem{liu2019clevr}
R.~Liu, C.~Liu, Y.~Bai, and A.~Yuille.
\newblock Clevr-ref+: Diagnosing visual reasoning with referring expressions.
\newblock In {\em Proceedings of the IEEE Conference on Computer Vision and
  Pattern Recognition (CVPR)}, 2019.

\bibitem{liu2018change}
Z.~Liu, G.~Li, G.~Mercier, Y.~He, and Q.~Pan.
\newblock Change detection in heterogenous remote sensing images via
  homogeneous pixel transformation.
\newblock {\em IEEE Transactions on Image Processing}, 27(4):1822--1834, 2018.

\bibitem{lu2017knowing}
J.~Lu, C.~Xiong, D.~Parikh, and R.~Socher.
\newblock Knowing when to look: Adaptive attention via a visual sentinel for
  image captioning.
\newblock In {\em Proceedings of the IEEE Conference on Computer Vision and
  Pattern Recognition (CVPR)}, 2017.

\bibitem{lu2016hierarchical}
J.~Lu, J.~Yang, D.~Batra, and D.~Parikh.
\newblock Hierarchical question-image co-attention for visual question
  answering.
\newblock In {\em Advances in Neural Information Processing Systems (NIPS)},
  2016.

\bibitem{lu2018neural}
J.~Lu, J.~Yang, D.~Batra, and D.~Parikh.
\newblock Neural baby talk.
\newblock In {\em Proceedings of the IEEE Conference on Computer Vision and
  Pattern Recognition (CVPR)}, pages 7219--7228, 2018.

\bibitem{luo2018discriminability}
R.~Luo, B.~Price, S.~Cohen, and G.~Shakhnarovich.
\newblock Discriminability objective for training descriptive captions.
\newblock In {\em Proceedings of the IEEE Conference on Computer Vision and
  Pattern Recognition (CVPR)}, 2018.

\bibitem{luo2017comprehension}
R.~Luo and G.~Shakhnarovich.
\newblock Comprehension-guided referring expressions.
\newblock In {\em Proceedings of the IEEE Conference on Computer Vision and
  Pattern Recognition (CVPR)}, 2017.

\bibitem{mascharka2018transparency}
D.~Mascharka, P.~Tran, R.~Soklaski, and A.~Majumdar.
\newblock Transparency by design: Closing the gap between performance and
  interpretability in visual reasoning.
\newblock In {\em Proceedings of the IEEE Conference on Computer Vision and
  Pattern Recognition}, 2018.

\bibitem{nam2016dual}
H.~Nam, J.-W. Ha, and J.~Kim.
\newblock Dual attention networks for multimodal reasoning and matching.
\newblock {\em arXiv:1611.00471}, 2016.

\bibitem{palazzolo2018fast}
E.~Palazzolo and C.~Stachniss.
\newblock Fast image-based geometric change detection given a 3d model.
\newblock In {\em 2018 IEEE International Conference on Robotics and Automation
  (ICRA)}, pages 6308--6315. IEEE, 2018.

\bibitem{papineni2002bleu}
K.~Papineni, S.~Roukos, T.~Ward, and W.-J. Zhu.
\newblock Bleu: a method for automatic evaluation of machine translation.
\newblock In {\em Proceedings of the Annual Meeting of the Association for
  Computational Linguistics (ACL)}, pages 311--318. Association for
  Computational Linguistics, 2002.

\bibitem{park2018multimodal}
D.~H. Park, L.~A. Hendricks, Z.~Akata, A.~Rohrbach, B.~Schiele, T.~Darrell, and
  M.~Rohrbach.
\newblock Multimodal explanations: Justifying decisions and pointing to the
  evidence.
\newblock In {\em Proceedings of the IEEE Conference on Computer Vision and
  Pattern Recognition (CVPR)}, 2018.

\bibitem{paszke2017automatic}
A.~Paszke, S.~Gross, S.~Chintala, G.~Chanan, E.~Yang, Z.~DeVito, Z.~Lin,
  A.~Desmaison, L.~Antiga, and A.~Lerer.
\newblock Automatic differentiation in pytorch.
\newblock In {\em Advances in Neural Information Processing Systems Workshops
  (NIPS Workshops)}, 2017.

\bibitem{patriarche2004review}
J.~Patriarche and B.~Erickson.
\newblock A review of the automated detection of change in serial imaging
  studies of the brain.
\newblock {\em Journal of digital imaging}, 17(3):158--174, 2004.

\bibitem{pedersoli2016areas}
M.~Pedersoli, T.~Lucas, C.~Schmid, and J.~Verbeek.
\newblock Areas of attention for image captioning.
\newblock In {\em Proceedings of the IEEE Conference on Computer Vision and
  Pattern Recognition (CVPR)}, 2016.

\bibitem{DBLP:journals/corr/abs-1709-07871}
E.~Perez, F.~Strub, H.~de~Vries, V.~Dumoulin, and A.~C. Courville.
\newblock Film: Visual reasoning with a general conditioning layer.
\newblock In {\em Proceedings of the Conference on Artificial Intelligence
  (AAAI)}, 2018.

\bibitem{radke2005image}
R.~J. Radke, S.~Andra, O.~Al-Kofahi, and B.~Roysam.
\newblock Image change detection algorithms: a systematic survey.
\newblock {\em IEEE Transactions on Image Processing}, 14(3):294--307, 2005.

\bibitem{sakurada2015change}
K.~Sakurada and T.~Okatani.
\newblock Change detection from a street image pair using cnn features and
  superpixel segmentation.
\newblock In {\em Proceedings of the British Machine Vision Conference (BMVC)},
  pages 61--1, 2015.

\bibitem{sakurada2017dense}
K.~Sakurada, W.~Wang, N.~Kawaguchi, and R.~Nakamura.
\newblock Dense optical flow based change detection network robust to
  difference of camera viewpoints.
\newblock {\em arXiv:1712.02941}, 2017.

\bibitem{stent2016precise}
S.~Stent, R.~Gherardi, B.~Stenger, and R.~Cipolla.
\newblock Precise deterministic change detection for smooth surfaces.
\newblock In {\em Proceedings of the IEEE Winter Conference on Applications of
  Computer Vision (WACV)}, pages 1--9. IEEE, 2016.

\bibitem{tian2014building}
J.~Tian, S.~Cui, and P.~Reinartz.
\newblock Building change detection based on satellite stereo imagery and
  digital surface models.
\newblock {\em IEEE Transactions on Geoscience and Remote Sensing},
  52(1):406--417, 2014.

\bibitem{vaduva2013latent}
C.~Vaduva, T.~Costachioiu, C.~Patrascu, I.~Gavat, V.~Lazarescu, and M.~Datcu.
\newblock A latent analysis of earth surface dynamic evolution using change map
  time series.
\newblock {\em IEEE Transactions on Geoscience and Remote Sensing},
  51(4):2105--2118, 2013.

\bibitem{vedantam2017context}
R.~Vedantam, S.~Bengio, K.~Murphy, D.~Parikh, and G.~Chechik.
\newblock Context-aware captions from context-agnostic supervision.
\newblock In {\em Proceedings of the IEEE Conference on Computer Vision and
  Pattern Recognition (CVPR)}, 2017.

\bibitem{vedantam2015cider}
R.~Vedantam, C.~Lawrence~Zitnick, and D.~Parikh.
\newblock Cider: Consensus-based image description evaluation.
\newblock In {\em Proceedings of the IEEE Conference on Computer Vision and
  Pattern Recognition (CVPR)}, 2015.

\bibitem{vinyals2015show}
O.~Vinyals, A.~Toshev, S.~Bengio, and D.~Erhan.
\newblock Show and tell: A neural image caption generator.
\newblock In {\em Proceedings of the IEEE Conference on Computer Vision and
  Pattern Recognition (CVPR)}, pages 3156--3164, 2015.

\bibitem{wang2014cdnet}
Y.~Wang, P.-M. Jodoin, F.~Porikli, J.~Konrad, Y.~Benezeth, and P.~Ishwar.
\newblock Cdnet 2014: An expanded change detection benchmark dataset.
\newblock In {\em Proceedings of the IEEE Conference on Computer Vision and
  Pattern Recognition Workshops (CVPR Workshops)}, pages 387--394, 2014.

\bibitem{xu2015show}
K.~Xu, J.~Ba, R.~Kiros, K.~Cho, A.~Courville, R.~Salakhudinov, R.~Zemel, and
  Y.~Bengio.
\newblock Show, attend and tell: Neural image caption generation with visual
  attention.
\newblock In {\em Proceedings of the International Conference on Machine
  Learning (ICML)}, pages 2048--2057, 2015.

\bibitem{yang2016stacked}
Z.~Yang, X.~He, J.~Gao, L.~Deng, and A.~Smola.
\newblock Stacked attention networks for image question answering.
\newblock In {\em Proceedings of the IEEE Conference on Computer Vision and
  Pattern Recognition (CVPR)}, pages 21--29, 2016.

\bibitem{yu2017joint}
L.~Yu, H.~Tan, M.~Bansal, and T.~L. Berg.
\newblock A joint speakerlistener-reinforcer model for referring expressions.
\newblock In {\em Proceedings of the IEEE Conference on Computer Vision and
  Pattern Recognition (CVPR)}, 2017.

\bibitem{zanetti2016generalized}
M.~Zanetti and L.~Bruzzone.
\newblock A generalized statistical model for binary change detection in
  multispectral images.
\newblock In {\em Geoscience and Remote Sensing Symposium (IGARSS), 2016 IEEE
  International}, pages 3378--3381. IEEE, 2016.

\bibitem{zhang2018top}
J.~Zhang, S.~A. Bargal, Z.~Lin, J.~Brandt, X.~Shen, and S.~Sclaroff.
\newblock Top-down neural attention by excitation backprop.
\newblock {\em International Journal of Computer Vision (IJCV)},
  126(10):1084--1102, 2018.

\end{thebibliography}
}

\end{document}